\documentclass[letterpaper]{article}
\usepackage{flairs}
\usepackage{times}
\usepackage{helvet}
\usepackage{courier}
\usepackage{graphicx}
\usepackage{natbib}
\usepackage{amsmath}
\begin{document}

\title{Explainable AI
in Deep Learning-Based Prediction of Solar Storms}
\author{Adam O. Rawashdeh, Jason T. L. Wang\\
Department of Computer Science\\
New Jersey Institute of Technology\\
Newark, NJ 07102, USA\\
\{aor9, wangj\}@njit.edu \\
\And Katherine G. Herbert\\
School of Computing \\
Montclair State University\\
Montclair, NJ 07043, USA\\
herbertk@montclair.edu\\
}

\maketitle

\begin{abstract}
\begin{quote}
A deep learning model is often considered a black-box model, as its internal workings tend to be opaque to the user. 
Because of the lack of transparency, it is challenging to understand the reasoning behind the model's predictions. 
Here, we present an approach to making
a deep learning-based solar storm prediction model interpretable,
where solar storms include solar flares and
coronal mass ejections (CMEs).
This deep learning model, built based on a long short-term memory (LSTM) network with an attention mechanism,
aims to predict whether an active region (AR) on the Sun's surface
that produces a flare within 24 hours will also
produce a CME associated with the flare.
 The crux of our approach is to model data samples in an AR as time series and use
the LSTM network to capture the temporal dynamics of the data samples.
To make the model's predictions accountable and reliable, 
we leverage post hoc model-agnostic techniques, which help elucidate the factors contributing to the predicted output for an input sequence
and provide insights into the model's behavior across multiple sequences within an AR.
To our knowledge, this is the first time that
interpretability has been added to an LSTM-based solar storm prediction model.
\end{quote}
\end{abstract}

\section{Introduction}
\label{sec:intro}
Solar storms include solar flares and coronal mass ejections (CMEs).
A solar flare is a large explosion in the Sun's atmosphere caused by tangling, crossing, or reorganizing of magnetic field lines.
CMEs are intense bursts of magnetic flux and plasma that are ejected from the Sun into interplanetary space
\citep{2020ApJ...890...12L}.
CMEs are often associated with solar flares and originate from active regions (ARs) in the Sun's photosphere, where magnetic fields are strong and evolve rapidly. Major CMEs and their associated flares can cause severe influences on the near-Earth environment, resulting in potentially life-threatening consequences \citep{2004SpWea...2.2004B}. Therefore, substantial efforts are being invested in the development of new technologies for the early detection and prediction of flares and CMEs
\citep{2016ApJ...821..127B,2018ApJ...861..128I}.

Both flares and CMEs are believed to be magnetically driven events; 
evidence shows that they may constitute different manifestations
of the same physical process
\citep{1995A&A...304..585H}. 
However, solar observations over the past few decades
have clearly indicated that there may not be a one-to-one correspondence
between flares and CMEs, and their relationship is still under active investigation. 
Much effort has been devoted to analyzing the structural properties of coronal magnetic fields, which may play an important role in determining whether an eruption evolves into a CME or remains as a confined flare. 
In the meantime, many researchers have investigated the relationship between CME productivity and the features of the photospheric magnetic field of flare-productive ARs where the features can be directly derived from photospheric vector magnetograms. 

For example,
\cite{2008automated} used properties such as flare duration
along with machine learning algorithms to predict whether a flare is likely to initiate a CME. 
\cite{2016ApJ...821..127B} used 
physical features, or predictive parameters, derived from the photospheric vector magnetic field data provided by the Helioseismic and Magnetic Imager (HMI)
on board the Solar Dynamics Observatory (SDO) to forecast whether a CME would be associated with a large flare. 
\cite{2018ApJ...861..128I} later developed methods to forecast whether flares would be associated with CMEs and solar energetic particles (SEPs). The authors used two machine learning algorithms, support vector machines (SVMs) and multilayer perceptrons (MLPs), and showed that SVMs performed slightly better than MLPs on the forecasting task. 
\cite{2020ApJ...890...12L}
designed a long short-term memory (LSTM) network 
and applied the LSTM network 
to SDO/HMI vector magnetic field data to predict whether an AR that produces a large flare within 24 hours will also produce a CME
associated with the flare. 

In this paper, we extend the work of
\cite{2020ApJ...890...12L}
by adding interpretability to the LSTM network using
two tools,
namely
SHapley Additive exPlanations (SHAP) and
Local Interpretable Model-agnostic Explanations (LIME)  
\citep{molnar2020interpretable}.
In what follows, we first
survey related work.
Then we present the results of SHAP,
followed by the results of LIME.
Finally, we conclude the paper and point out some directions for future work.

\section{Related Work}

Interpretable machine learning, or explainable artificial intelligence (XAI), has recently drawn much attention in the AI field
as researchers and practitioners seek to provide more transparency to their models. 
\cite{miller2019explanation} defined the interpretability of a model as
\textit{the degree to which an observer can understand the cause of a decision}.
The terms ``interpretable'' and ``explainable'' 
are often used interchangeably, and the basic idea behind them is the same.
One may wonder, when a machine learning model
gives a prediction with high accuracy, 
why should we bother about the reason behind this prediction?
The answer to that question is that
it depends on what we are trying to predict. 
If the prediction task is simple and the only thing we are concerned about is the performance, then explanations may not be needed.
On the other hand, for certain complex prediction tasks, 
 it is very important and valuable to understand the reason behind a prediction and to be able to explain the prediction \citep{molnar2020interpretable}.

Recently, XAI has been incorporated into
solar flare predictions.
For example, \cite{feldhaus2021explainable} implemented LIME in a solar flare prediction model built using SVMs. 
\cite{li2022fast} developed a model named FAST-CF, which provides intuitive post-hoc counterfactual explanations for
solar flare predictions. 
\cite{pandey2023explainable} performed a post hoc analysis on a deep learning-based full-disk solar flare prediction model using three methods, including class activation mapping, deep SHAP, and integrated gradients.
In contrast to the above studies, 
our work focuses on
incorporating
SHAP and LIME into
an LSTM network for
solar storm predictions using photospheric magnetic field parameters. 

\section{Preliminaries}

\subsection{XAI Background}

XAI consists of a set of tools and libraries designed to explain predictions for machine learning algorithms. It is commonly used when dealing
with black-box models such as LSTM. Its importance lies in the opaque nature of the black-box models, where only the inputs and outputs are known. However, their inner workings remain enigmatic. 
Incorporating XAI is a key factor in the deployment of responsible AI. 
XAI helps to ensure that the results of a model can be explained and understood. Adding interpretability to a black-box model will result in beneficial outcomes such as mitigating risk, reducing bias, and increasing trust.  
 
We propose to incorporate
SHAP and LIME into
the LSTM model that originates from
\cite{2020ApJ...890...12L} for
solar storm predictions.
SHAP handles global explanations, while
LIME is dedicated to local explanations.
Global explanations are used to understand the model as a whole
on the basis of the entire test set.
It is useful to see which feature plays the biggest role in the model's output. Local explanations, on the other hand, take a look at specific instances
in the test set. 
Through local explanations, we can better understand each 
feature and why certain behaviors lead to specific predictions.
For both SHAP and LIME, we adopt a random seed function.
Specifically, we use np.random.seed(42),
where the seed number, 42, is commonly used
in the machine learning community.

\subsection{Data and Feature Space}

This study included 33,604 data samples in the training set and 540 data samples in the test set. The dataset size was predetermined by a forecast horizon (24 hrs) to predict whether a 
solar active region (AR) that produces a flare within 24 hrs will also produce a CME associated with the flare.

We used 12 magnetic field parameters or features, listed in Table \ref{features}, from
the Space-weather HMI Active Region Patches 
\citep[SHARPs;][]{bobra2014} 
to make the predictions.
Each data sample contains the 12 features, along with a label
of ``P" indicating that the data sample is in the positive class (i.e., the corresponding AR that produces a flare within 24 hours will produce a CME associated with the flare), 
or a label of ``N" indicating that the data sample is in the negative class (i.e., the corresponding AR that produces a flare within 24 hours will not produce a CME associated with the flare).
The 12 features in Table \ref{features} produced the highest mean True Skill Statistics (TSS),
which was 0.562,
for the 24-hour forecast horizon
\citep{2020ApJ...890...12L}. 

\begin{table}[ht]
\caption{Features Used for Solar Storm  Predictions}
\label{features}
\begin{center}
\begin{tabular}{|p{2cm}|p{5.6cm}|}
    \hline
    Feature & Description \\
    \hline
    TOTUSJZ & Total unsigned vertical current \\
    USFLUX & Total unsigned flux \\
    TOTPOT & Total magnetic free energy density \\
    SAVNCPP & Sum of the net current per polarity\\
    ABSNJZH & Absolute value of net current helicity  \\
    MEANPOT & Mean magnetic free energy  \\
    MEANSHR & Mean shear angle  \\
    SHRGT45 & Area fraction with shear
    $\geq$ 45$^{\circ}$ \\
    MEANJZH & Mean current helicity \\
    MEANGAM & Mean angle of field from radial \\
    MEANALP & Mean characteristic twist parameter \\
    MEANGBZ & Mean gradient of vertical field \\
    \hline
\end{tabular}
\end{center}
\end{table}

\section{Global Explanations with SHAP}
\label{sec:SHAP}

\subsection{Feature Analysis}

\begin{figure}[ht]
    \centering
    \includegraphics[width=\linewidth]{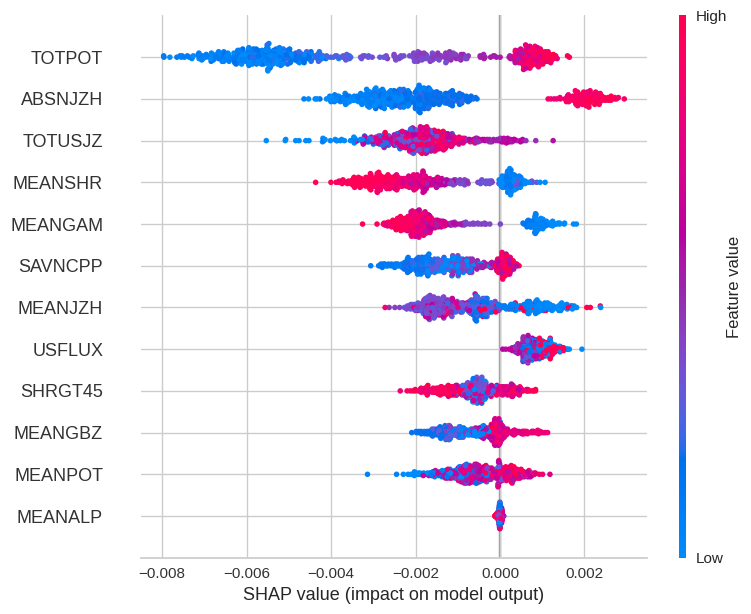}
    \caption{Beeswarm plot to visualize the positive or negative effect of a feature 
     for each test sample, represented by a color dot, on the model's predictions.}
    \label{SHAP_Beeswarm}    
\end{figure}
  
SHAP incorporates game theory to
give each feature a SHAP value.
Positive SHAP values have a positive effect, 
while negative SHAP values
have a negative effect. 
A positive effect increases the probability of
predicting that a test sample is in the positive class,
whereas a negative effect
increases the probability of predicting that a test sample is in the negative class. 
We adopted shap.GradientExplainer in this study.

Figure \ref{SHAP_Beeswarm} presents the beeswarm plot for the LSTM model.
In the beeswarm plot, we can observe, for each feature, the distribution of SHAP values for all test samples. 
Each test sample corresponds to a dot in each feature row. 
The placement of each dot on the X-axis is determined by the SHAP value that the corresponding test sample receives. 
When dots cluster,
it shows common test samples based on their SHAP values. The color of a dot depends on
the value of a feature in the corresponding test sample. 
Red indicates a high feature value, blue indicates a low feature value, and purple indicates an average or moderate feature value.

Now, focus on the TOTPOT feature in Figure \ref{SHAP_Beeswarm}.
There are more test samples with negative SHAP values than with positive SHAP values.
This means that the TOTPOT feature is
more likely to push the model's predictions towards a negative class.
Furthermore, 
low feature values (blue dots) tend
to lead to negative predictions, 
as these low feature values have negative SHAP values.
High feature values (red dots) tend to lead to positive predictions, as these high feature values have positive SHAP values.

Next, focus on the MEANALP feature in Figure \ref{SHAP_Beeswarm}. 
If we compare the MEANALP feature with the TOTPOT feature,
MEANALP may appear to have fewer dots. 
The reason for this appearance is that most of the SHAP values for MEANALP are zero
or near zero. 
This causes the dots to cluster up and makes MEANALP seem to have fewer dots.
Furthermore, since most of the SHAP values for MEANALP are zero or near zero,
the MEANALP feature neither pushes the model's predictions towards a negative class
nor pushes the model's predictions towards a positive class,
implying that the MEANALP feature plays an unimportant role in
the model's predictions. 

\begin{figure}[ht]
    \centering
    \includegraphics[width=\linewidth]{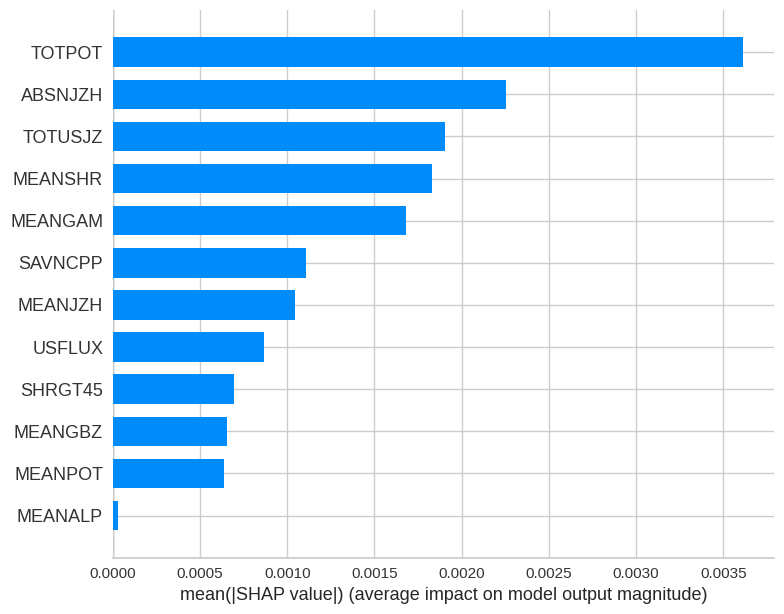}
    \caption{Bar plot to display the global importance of each feature on the model's predictions.}
    \label{SHAP_Bar}    
\end{figure} 

Figure \ref{SHAP_Bar} presents the bar plot for 
the LSTM model. 
In the bar plot, each feature is given
the mean of the absolute SHAP values 
across all test samples.
This mean of the absolute SHAP values
is how the importance of each feature is measured in the bar plot.
The longer the bar in the bar plot, the more important the corresponding feature is to the model's predictions. 
We can see in Figure \ref{SHAP_Bar} that the TOTPOT feature is of the highest importance to the model's predictions, while the MEANALP feature is of the lowest importance to the model's predictions.

In contrast to the bar plot in Figure \ref{SHAP_Bar},
the beeswarm plot in Figure \ref{SHAP_Beeswarm} 
displays a separate SHAP value for each test sample, and
shows the variability of the test samples, represented by color dots, for each feature. The fewer clusters and the more spread out from the zero
in the beeswarm plot
indicate higher SHAP values (positive or negative), causing a higher mean of absolute SHAP values and ultimately a higher importance.
Now, consider again the appearance of MEANALP in the beeswarm plot in Figure \ref{SHAP_Beeswarm}. The result of having most of its SHAP values equal or near zero 
reflects how short its bar is in the bar plot
in Figure \ref{SHAP_Bar}.
On the other hand, the nature of high variability
of the TOTPOT feature in the beeswarm plot in Figure \ref{SHAP_Beeswarm} causes its bar to
be significantly longer than those of the other features
in the bar plot in Figure \ref{SHAP_Bar}.

\begin{figure}[t]
    \centering
    \includegraphics[width=\linewidth]{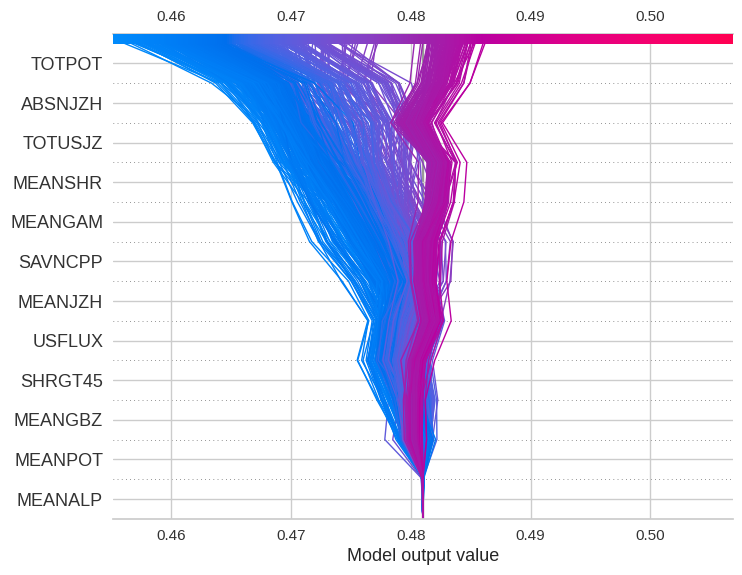}
    \caption{Decision plot to understand 
    how the model produces its predictions.}
    \label{SHAP_Decision}    
\end{figure}

Figure \ref{SHAP_Decision} presents the decision plot for the LSTM model. On the Y-axis of the decision plot, the features are displayed, from top to bottom, based on their importance, with the most important feature displayed at the top and the least important feature displayed at the bottom.
Each prediction/test sample is represented by a line
in the decision plot. 
The predictions for the test samples begin at the bottom with the same base value, which is approximately 0.48. This base value is the mean of the model's prediction values across all data samples 
in the training set. 
It is used as a starting point before considering any feature contribution. 
 As a prediction/line moves from bottom to top, the SHAP value for each feature is added to the base value. This can help to understand the contribution of each feature in the prediction. 
The features can have positive or negative contributions, pushing the
corresponding line to the right or left, respectively.

The prediction value (i.e., the model output value)
of a test sample
is the probability that the test sample belongs to
the positive class.
For example, a prediction value of 0.4 indicates that there is a 40\% chance that the test sample belongs to the positive class.
At the top, each prediction/line reaches its prediction value. The prediction value determines the corresponding line color of the prediction.
A blue line indicates a low prediction value, while a red line indicates a high prediction value with respect to the base value.
Specifically, the features in blue reduce the model output value compared to the base value.
The deeper the blue, the stronger the negative contribution of that feature.
The features in red increase the model's output value compared to the base value.
The deeper the red, the stronger the positive contribution of that feature.

\begin{figure*}[ht]
    \centering
    \includegraphics[width=0.7\linewidth]{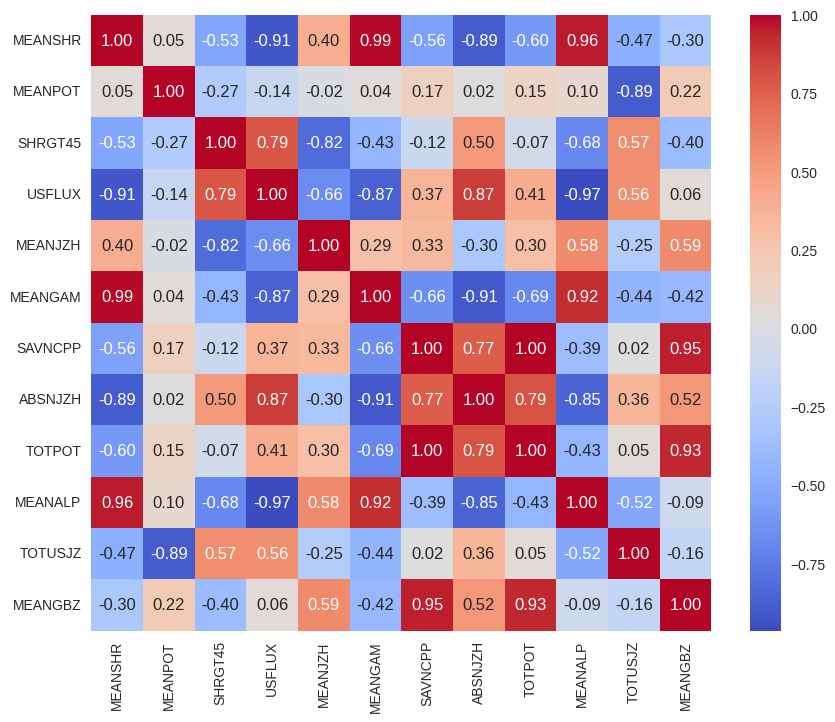}
    \caption{Pearson's correlation coefficient matrix for the 12 features used in this study.}
    \label{PearsonCorrelationMatrix}    
\end{figure*}

The results shown in the beeswarm, bar, and decision plots are consistent. 
Based on these results, 
we conclude that the TOTPOT feature is the most important feature in the model's predictions, while the MEANALP feature has the least importance in the model's predictions. Other features such as MEANPOT and MEANGBZ are indicated as features of low importance, while features such as ABSNJZH and TOTUSJZ are of high importance.

\begin{figure}[ht]
    \centering
    \includegraphics[width=\linewidth]{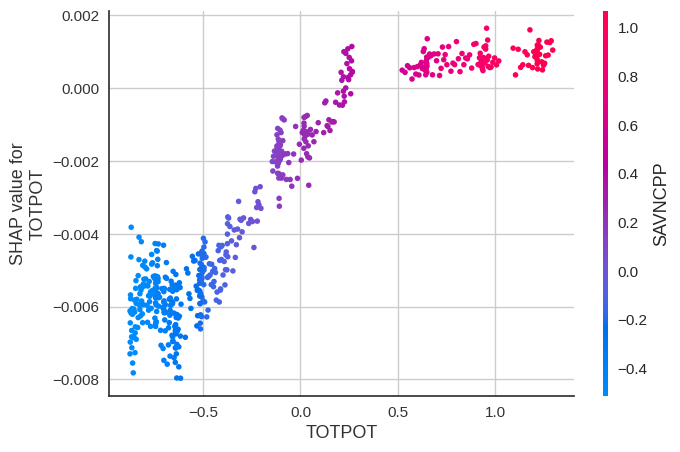}
    \caption{Dependence plot for the most important feature, TOTPOT, with its strongest correlate, SAVNCPP.}
    \label{PDP_PositiveStrong}
\end{figure}

\begin{figure}[ht]
    \centering
    \includegraphics[width=\linewidth]{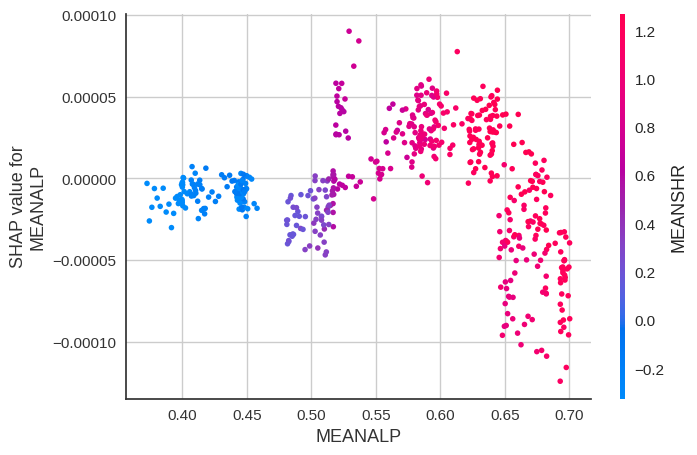}
    \caption{Dependence plot for the least important feature, MEANALP, with its strongest correlate, MEANSHR.}
    \label{PDP_LeastStrong}
\end{figure}

\subsection{Feature Interaction Analysis}

Figure \ref{PearsonCorrelationMatrix} presents the Pearson correlation
coefficient matrix for the 12 features used in this study.
The matrix shows the correlation between all two features. The correlation score is valued in the range $[-1, 1]$. A score near 1 indicates a high positive correlation, a score near $-1$ indicates a high negative correlation, and a score near 0 indicates little or no correlation. 

It can be seen in Figure \ref{PearsonCorrelationMatrix} that
the most important feature,
TOTPOT, and the SAVNCPP feature have a correlation value of 1, the highest possible value indicating a strong correlation. TOTPOT and TOTUSJZ have a correlation value of 0.05, a value very close to 0 indicating a very weak correlation. TOTPOT and MENGAM have a correlation value of $-0.69$, a moderately low value indicating a negative correlation. 
It can also be seen in Figure \ref{PearsonCorrelationMatrix}
that the least important feature, MEANALP, and the MEANSHR feature have a correlation value of 0.96, a high value indicating a strong correlation. MEANALP and MEANGBZ have a correlation value of $-0.09$, a value very close to 0 indicating a very weak correlation. MEANALP and USFLUX have a correlation value of $-0.97$, a low value indicating a negative correlation.

Now, consider the most important feature, TOTPOT.
As mentioned above, the feature that has the highest positive correlation with
TOTPOT is SAVNCPP, with a Pearson correlation coefficient of 1.
Figure \ref{PDP_PositiveStrong} presents the dependence plot
between TOTPOT and SAVNCPP.
 The X-axis of the dependence plot represents the TOTPOT feature values, where the feature values have been normalized \citep{2020ApJ...890...12L}. 
 The Y-axis shows the SHAP values of TOTPOT.
 The color bar on the right-hand side displays the normalized SAVNCPP feature values, where the large/positive SAVNCPP feature values are in red, while the small/negative SAVNCPP feature values are in blue.
 Each dot in the dependence plot represents a test sample.
The dependence plot shows the effects of the interaction between
TOTPOT and SAVNCPP.
We see that a test sample with a small/negative TOTPOT feature value
and a small/negative (blue) SAVNCPP feature value has a negative SHAP value,
indicating that the test sample is
likely to be negative.
However, a test sample with a large/positive TOTPOT feature
value and a large/positive (red) SAVNCPP feature value has a positive
SHAP value, indicating that the test sample is 
likely to be positive.
There are more test samples 
likely to be in the negative class (with negative SHAP values) than in the positive class (with positive SHAP values).

Next, consider the least important feature, MEANALP. The feature that has the highest positive correlation with MEANALP is MEANSHR, with a Pearson correlation coefficient of 0.96. 
Figure \ref{PDP_LeastStrong} presents the dependence plot between MEANALP and MEANSHR, showing the effects of the interaction between MEANALP and MEANSHR. 
It can be seen in Figure \ref{PDP_LeastStrong} that a test sample with a small MEANALP feature value and a small/negative (blue) MEANSHR feature value has a negative SHAP value, indicating that the test sample is likely to be in the negative class.
There are more test samples likely
to be in the negative class (with negative SHAP values) than in the positive class (with positive SHAP values).

\section{Local Explanations with LIME}
\label{sec:LIME}

Local explanations look at individual predictions 
made by the LSTM model. 
With LIME, we can better understand each feature and why certain behaviors lead to a specific prediction.
Another characteristic of LIME is that it is model-agnostic. This means that it can be applied to any machine learning model dealing with any input data such as text, image, or, in the case of our study, multivariate time series. 
The flexibility of the LIME tool allows it to be implemented no matter the complexity of the inner workings of a deep learning model. 
We adopted lime.lime\_tabular.LimeTabularExplainer in this study.

\begin{figure}[ht]
    \centering
    \includegraphics[width=\linewidth]{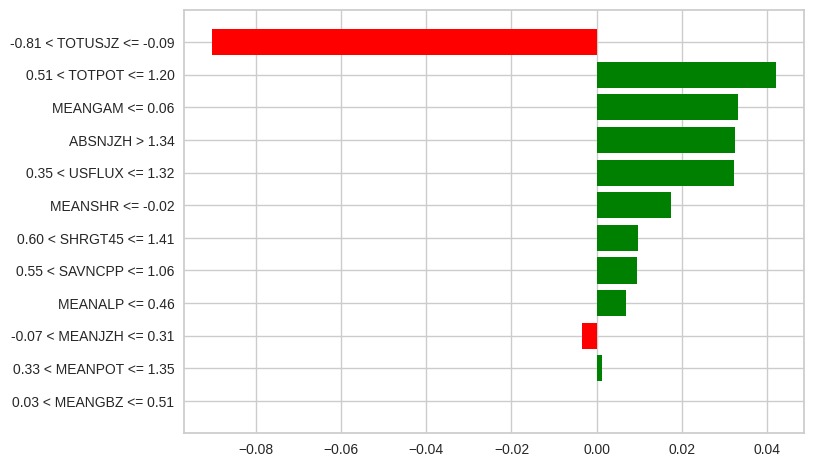}
    \caption{LIME plot for a 
    test sample predicted to be in the positive class.}
    \label{LIME_pos}
\end{figure}

\begin{figure}[ht]
    \centering
    \includegraphics[width=\linewidth]{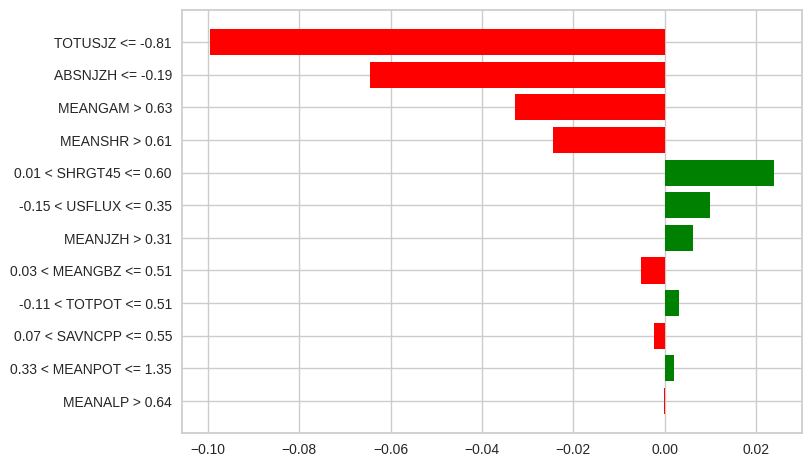}
    \caption{LIME plot for 
    a test sample predicted to be in the negative class.}
    \label{LIME_neg}
\end{figure}

Figure \ref{LIME_pos} 
presents the LIME plot for a test sample predicted to be in the positive class.
Figure \ref{LIME_neg}
presents the LIME plot for a test sample predicted to be in the negative class.
The X-axis shows LIME values.
The features on the Y-axis 
are ranked from highest to lowest importance.
We need to distinguish the negative and positive bars with respect to the LIME values on the X-axis. When a feature favors one side over the other, it means that the model is identifying the class to which it believes the test sample should belong. 
Negative LIME values fall into the negative class, while positive LIME values fall into the positive class.

Consider, for example, Figure \ref{LIME_pos}.
Most of the features in the figure have positive bars
and the positive magnitude outweighs the negative magnitude.
Thus, the model predicts that the test sample
in Figure \ref{LIME_pos} is in the positive class.
On the other hand, in Figure \ref{LIME_neg},
most of the features have negative bars
and the negative magnitude outweighs the positive magnitude, which strongly contributes to the negative class.
Consequently, the model predicts that the test sample
in Figure \ref{LIME_neg}
is in the negative class.

In both Figures \ref{LIME_pos} and \ref{LIME_neg}, each feature on the Y-axis is associated with a range. 
For example, in Figure \ref{LIME_pos}, ABSNJZH with a positive bar has a range greater than $1.34$; this means that ABSNJZH values greater than $1.34$ increase the chances that a test sample belongs to the positive class.
On the other hand, in Figure \ref{LIME_neg}, TOTUSJZ with a negative bar has a range less than or equal to $-0.81$; this means that TOTUSJZ values less than or equal to $-0.81$ decrease the chances that a test sample belongs to the positive class (or increase the chances that the test sample belongs to the negative class).

When comparing the LIME plots to the SHAP plots, it is important to recall that SHAP is designed for
global explanations considering
all data samples in the test set,
while LIME is designed for local explanations inspecting specific test samples. However, comparing the two plots can provide valuable insight to identify possible overlaps in importance of the features. 
TOTUSJZ, for example, is of high importance in both LIME plots and SHAP plots.

\section{Conclusion}
\label{sec:conclusion}

In this paper, we present an interpretable LSTM model for solar storm predictions
by incorporating two XAI tools (SHAP and LIME) into the model. The SHAP tool helps us understand the features that have the largest impact on the overall predictions of the model. The LIME tool helps us explore and explain the individual predictions made by the model. 
In addition, we perform feature interaction analysis using 
Pearson's correlation coefficient matrix and
dependence plots to visualize interaction effects between two features. 
In the future we plan to extend our work to build interpretable deep learning models
for other space weather events such as
solar energetic particles, interplanetary shocks and geomagnetic storms.

\bibliographystyle{flairs}

\begin{thebibliography}{}

\bibitem[\protect\citeauthoryear{Baker \bgroup et al\mbox.\egroup }{2004}]{2004SpWea...2.2004B}
Baker, D.~N.; Daly, E.; Daglis, I.; Kappenman, J.~G.; and Panasyuk, M.
\newblock 2004.
\newblock Effects of space weather on technology infrastructure.
\newblock {\em Space Weather} 2:S02004.

\bibitem[\protect\citeauthoryear{Bobra and Ilonidis}{2016}]{2016ApJ...821..127B}
Bobra, M.~G., and Ilonidis, S.
\newblock 2016.
\newblock Predicting coronal mass ejections using machine learning methods.
\newblock {\em The Astrophysical Journal} 821:127.

\bibitem[\protect\citeauthoryear{Bobra \bgroup et al\mbox.\egroup }{2014}]{bobra2014}
Bobra, M.~G.; Sun, X.; Hoeksema, J.~T.; and et~al.
\newblock 2014.
\newblock {The Helioseismic and Magnetic Imager (HMI)} vector magnetic field pipeline: {SHARPs} - space-weather {HMI} active region patches.
\newblock {\em Solar Physics} 289(11):3549--3578.

\bibitem[\protect\citeauthoryear{Feldhaus and Carande}{2021}]{feldhaus2021explainable}
Feldhaus, C., and Carande, W.
\newblock 2021.
\newblock Explainable artificial intelligence for solar flare prediction.
\newblock In {\em AGU Fall Meeting Abstracts}, volume 2021,  NG45B--0576.

\bibitem[\protect\citeauthoryear{Harrison}{1995}]{1995A&A...304..585H}
Harrison, R.~A.
\newblock 1995.
\newblock {The nature of solar flares associated with coronal mass ejection.}
\newblock {\em Astronomy \& Astrophysics} 304:585.

\bibitem[\protect\citeauthoryear{Inceoglu \bgroup et al\mbox.\egroup }{2018}]{2018ApJ...861..128I}
Inceoglu, F.; Jeppesen, J.~H.; Kongstad, P.; Hernández~Marcano, N.~J.; Jacobsen, R.~H.; and Karoff, C.
\newblock 2018.
\newblock Using machine learning methods to forecast if solar flares will be associated with {CMEs} and {SEPs}.
\newblock {\em The Astrophysical Journal} 861:128.

\bibitem[\protect\citeauthoryear{Li \bgroup et al\mbox.\egroup }{2022}]{li2022fast}
Li, P.; Bahri, O.; Boubrahimi, S.~F.; and Hamdi, S.~M.
\newblock 2022.
\newblock Fast counterfactual explanation for solar flare prediction.
\newblock In {\em Proceedings of the 21st IEEE International Conference on Machine Learning and Applications},  1238--1243.

\bibitem[\protect\citeauthoryear{{Liu} \bgroup et al\mbox.\egroup }{2020}]{2020ApJ...890...12L}
{Liu}, H.; {Liu}, C.; {Wang}, J. T.~L.; and {Wang}, H.
\newblock 2020.
\newblock Predicting coronal mass ejections using {SDO/HMI} vector magnetic data products and recurrent neural networks.
\newblock {\em The Astrophysical Journal} 890(1):12.

\bibitem[\protect\citeauthoryear{Miller}{2019}]{miller2019explanation}
Miller, T.
\newblock 2019.
\newblock Explanation in artificial intelligence: Insights from the social sciences.
\newblock {\em Artificial intelligence} 267:1--38.

\bibitem[\protect\citeauthoryear{Molnar}{2020}]{molnar2020interpretable}
Molnar, C.
\newblock 2020.
\newblock {\em Interpretable Machine Learning}.
\newblock lulu.com.

\bibitem[\protect\citeauthoryear{Pandey \bgroup et al\mbox.\egroup }{2023}]{pandey2023explainable}
Pandey, C.; Angryk, R.~A.; Georgoulis, M.~K.; and Aydin, B.
\newblock 2023.
\newblock Explainable deep learning-based solar flare prediction with post hoc attention for operational forecasting.
\newblock In {\em Proceedings of the 2023 International Conference on Discovery Science},  567--581.

\bibitem[\protect\citeauthoryear{Qahwaji \bgroup et al\mbox.\egroup }{2008}]{2008automated}
Qahwaji, R.; Colak, T.; Al-Omari, M.; and Ipson, S.
\newblock 2008.
\newblock Automated prediction of {CMEs} using machine learning of {CME}–flare associations.
\newblock In Ireland, J., and Young, C.~A., eds., {\em Solar Image Analysis and Visualization}. New York: Springer.
\newblock  261.

\end{thebibliography}

\end{document}